\pdfoutput=1
\documentclass[11pt]{article}

\usepackage[preprint]{acl}

\usepackage{times}
\usepackage{latexsym}
\usepackage{microtype}
\usepackage{graphicx}
\usepackage{subfigure}
\usepackage{booktabs} % for professional tables
\usepackage{bbm}
\usepackage[T1]{fontenc}
\usepackage{algorithm}
\usepackage{algorithmic}
\usepackage{amsmath}
\DeclareMathOperator*{\argmax}{arg\,max} % thin space, limits underneath in displays
\usepackage{tcolorbox}

\makeatletter
\newcommand*\footnotescript{%
  \@setfontsize\footnotescript{8.5}{9.5}%
}
\makeatother

\makeatletter
\newcommand*\smallnormalsize{%
  \@setfontsize\smallnormalsize{10.8}{11.8}%
}
\makeatother

\makeatletter
\newcommand*\smallnormallsize{%
  \@setfontsize\smallnormallsize{10}{11}%
}
\makeatother

\makeatletter
\newcommand*\smallnormalllsize{%
  \@setfontsize\smallnormallsize{11.5}{12.5}%
}
\makeatother

\usepackage{arydshln}

\usepackage[utf8]{inputenc}

\usepackage{microtype}

\usepackage{inconsolata}

\makeatletter
\newcommand{\verbatimfont}[1]{\def\verbatim@font{#1}}%
\makeatother

\NewDocumentCommand{\arafat}
{ mO{} }{\textcolor{blue}{\textsuperscript{\textit{Arafat}}\textsf{\textbf{\small[#1]}}}}

\newcommand\blfootnote[1]{
    \begingroup
    \renewcommand\thefootnote{}\footnote{#1}
    \addtocounter{footnote}{-1}
    \endgroup
}

\title{%Content-Grounded %Principle-driven
Self-Refinement of Language Models %with 
from \\ External Proxy Metrics Feedback}

\author{\smallnormalllsize{\hspace{4.5mm}\textbf{Keshav Ramji$^{\hspace{0.5mm} 1 \hspace{0.5mm} \footnotemark[0 *]}$ \hspace{2mm} Young-Suk Lee$^{\hspace{0.5mm} 2}$ \hspace{2mm} Ramón Fernandez Astudillo$^{\hspace{0.5mm} 2}$\hspace{2mm} Md Arafat Sultan$^{\hspace{0.5mm} 2}$}}\\ \smallnormalllsize{\textbf{Tahira Naseem$^{\hspace{0.5mm} 2}$ \hspace{2mm} Asim Munawar$^{\hspace{0.5mm} 2}$ \hspace{2mm} Radu Florian$^{\hspace{0.5mm} 2}$ \hspace{2mm} Salim Roukos$^{\hspace{0.5mm} 2}$}} \\
  \smallnormalllsize{${}^{1} \hspace{-0.5mm}$ University of Pennsylvania \hspace{7.5mm} 
  ${}^{2 \hspace{-0.5mm}}$ IBM Research AI} \\
  \smallnormallsize{
  \texttt{keshavr@seas.upenn.edu}} \\
  \smallnormallsize{\texttt{\{ysuklee, tnaseem, raduf, roukos\}@us.ibm.com}} \\
  \smallnormallsize{\texttt{\{ramon.astudillo, arafat.sultan, asim\}@ibm.com}}}

\begin{document}

\maketitle
\begin{abstract}
%It is desirable for Large Language Models (LLMs) to capture multiple objectives when providing a response, just as humans would.
It is often desirable for Large Language Models (LLMs) to capture multiple objectives when providing a response.
In document-grounded response generation, for example, agent responses are expected to be \textit{relevant} to a user's query while also being \textit{grounded} in a given document.
%In a similar vein to how one assesses their responses heuristically along several dimensions in their mental model, we introduce \_\_\_\_ to enable self-refinement on these principles to yield higher-quality outputs.
%In this paper, we introduce \_\_\_\_, which enables an LLM to refine its own initial response along key dimensions of quality -- also referred to as \textit{\mbox{principles}} -- yielding an overall better final response.
In this paper, we introduce \textbf{Pro}xy \textbf{M}etr\textbf{i}c-based \textbf{Se}lf-Refinement (\textbf{ProMiSe}), which enables an LLM to refine its own initial response along key dimensions of quality guided by external metrics feedback, yielding an overall better final response.
%The algorithm first generates an initial response from the LLM while applying rejection sampling, then obtains implicit external feedback on response quality through inexpensive, task-specific proxy metrics, and iteratively refines the response with respect to the desired principles.
ProMiSe leverages feedback on response quality through principle-specific proxy metrics, and iteratively refines its response one principle at a time.
%In particular, the ensemble of scoring feedback mechanisms renders this method an implicit, parameter-free form of reinforcement learning.
%Our proposed algorithm works either zero-shot or in-context: response generation and principle-driven refinement are performed using few-shot exemplars, making the approach domain-agnostic. 
%We apply \_\_\_\_ to conversational question answering from documents, demonstrating its effectiveness in enabling self-refinement with open-source and smaller language models than previously shown (non-GPT) via in-context learning, further evidenced when fine-tuning on synthetic multi-turn conversational data generated by the algorithm.
We apply ProMiSe to open source language models \textsc{flan-t5-xxl} and \textsc{llama-2-13b-chat}, to evaluate its performance on document-grounded question answering datasets, MultiDoc2Dial and QuAC, 
demonstrating that self-refinement improves response quality.
%observing significant performance improvements.
%We apply ProMiSe to document-grounded conversational question answering datasets, %demonstrating its domain-agnostic nature across the 
%MultiDoc2Dial and QuAC, observing performance improvement by a significant margin. We also restrict models used to open source models, such as \textsc{flan-t5-xxl} and \textsc{llama-2-13b-chat} that are smaller than those used in prior work. 
We further show that fine-tuning \textsc{llama-2-13b-chat} on the synthetic dialogue data generated by ProMiSe yields significant performance improvements over the zero-shot baseline as well as a supervised fine-tuned model on human annotated data.
%Further evidence is provided via model fine-tuning on synthetic multi-turn conversational data generated by the algorithm, yielding improvements of +6-6.5\% on Recall measures and +7.7\% on K-Precision measures.
%\arafat{We should provide actual numbers here when all the results are back...}
\end{abstract}

\section{Introduction}
\label{introduction}

The state-of-the-art large language models (LLMs) have demonstrated to be effective in generating new synthetic data, useful in improving zero-shot task generalization through fine-tuning without requiring vast amounts of human annotations. Various approaches have been proposed to show the ability of models to evaluate and critique responses \cite{saunders2022selfcritiquing, scheurer2023training, shinn2023reflexion, selfee2023},
as well as their potential to refine: given feedback, correct their outputs \cite{welleck2022generating, peng2023check, madaan2023selfrefine, huang2023large, wang2023enable}. These explorations have studied various feedback mechanisms (human-in-the-loop, reward models to capture human preferences, model-generated feedback) and forms (pairwise comparisons, scalar scores, natural language descriptions), as well as refinement techniques (separate supervised refiners, domain-specific refinement). 

\blfootnote{${}^{*}$ Work done during internship at IBM Research AI.}
Of particular note are recent works exploring the self-refinement phenomenon \cite{madaan2023selfrefine, wang2023enable, shinn2023reflexion}, leveraging the same LLM to perform critique and/or refinement on top of generating responses. The observations of these works unveil shortcomings: smaller instruction-tuned models fail to replicate the results of systems such as GPT-3.5 and GPT-4 in refinement, and in the absence of well-designed stopping mechanisms, 
%self-correction can overoptimize; that is, 
self-refinement applied to high-quality responses can make the results worse \cite{huang2023large}. When humans correct themselves, they do it often with one or more objectives in mind, i.e. principles. 
%(henceforth referred to as principles). 
Such principles may include faithfulness, specificity, safety (i.e. non-toxic), relevance to a question posed, etc. and may vary across tasks — we seek to imbue these aspects into conversational agents, to ensure they are reflected in the agent's responses.

\begin{figure*}[t]
    \centering
    \includegraphics[width=\textwidth]{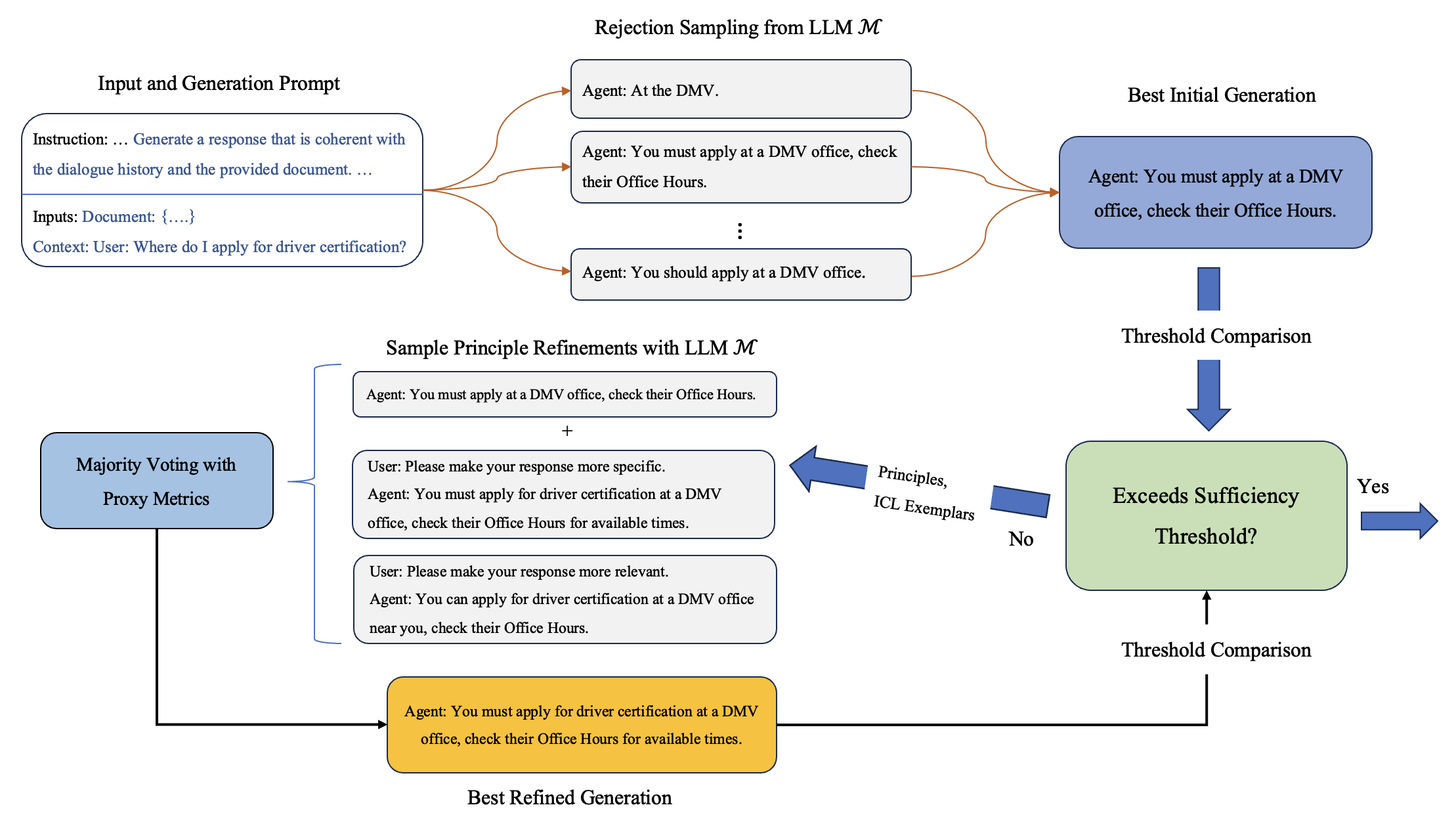}
    \vspace{-6mm}
    \caption{A high-level overview of our proposed self-refinement algorithm for content-grounded question answering, with both initial response generation and iterative refinement performed with the same Large Language Model $\mathcal{M}$. %Given an input, we provide the initial generation prompt with in-context exemplars, perform rejection sampling and determine the sufficiency of the best response thus far. If the response is deemed insufficient, then we proceed to sample refinements based on our chosen principles, perform majority voting to determine a "centroid" response, and then check the sufficiency of the best refinment. If a response meets the threshold, we accept it. 
    }
\end{figure*}

To this effect, we introduce an iterative, principle-guided approach to self-refinement in relatively smaller language models where refinement has previously proven unsuccessful. Our algorithm, termed \textbf{Pro}xy \textbf{M}etr\textbf{i}c-based \textbf{Se}lf-Refinement %(\textbf{ProMiSe}), seeks to leverage inexpensive sources of feedback through deliberately chosen, task-appropriate metrics with calibrated sufficiency thresholds. We employ few-shot demonstrations of refinement on a particular principle, yielding prompts better designed to match the instruction-following capabilities of said smaller models, as opposed to simultaneous refinement on many dimensions. 
(\textbf{ProMiSe}), combines proxy metric thresholding for different principles with independent principle-specific few-shot refinement and best-of-N rejection sampling. %Our approach samples diverse candidate responses for a given input, as well as for each refinement iteration, and uses the proxy metrics to determine the most promising candidate. 
This allows for the deliberate selection of task-appropriate metrics with calibrated sufficiency thresholds, and specific prompts better designed to match the instruction-following capabilities of smaller models. In this manner, we perform multi-aspect self-refinement via iterative single-aspect improvement queries, as opposed to simultaneous refinement on many dimensions. %in each phase. 

We apply this method to content-grounded question answering, demonstrating consistent improvements on a diverse set of evaluation metrics for single-turn response generation.
%both lexically and semantically-oriented metrics, in an entirely few-shot manner for single-turn response generation. 
We then extend ProMiSe to  multi-turn dialogue data generation to generate user queries in addition to agent responses. We fine-tune \textsc{llama-2-13b-chat} on the synthetic data, yielding significant improvement over the zero-shot baseline and supervised models solely fined-tuned on  human annotations.  
%yielding increases of 6-8\% across all metrics.

Crucially, this approach is built on open-source models and does not rely on propietary models with black-box API access; we note, however, that the proposed algorithm can be directly applied to closed-source models as well. Furthermore, it can be extended to other tasks, provided that proxy metrics can be defined and a few in-context exemplars can be created for the relevant principles.  
Our key contributions are:

\begin{itemize}
\vspace{-1.25mm}
\item We introduce a novel domain-agnostic algorithm, \textbf{ProMiSe}, to perform multi-aspect self-refinement on desirable principles for a response through in-context learning, using proxy metrics as external %response 
quality feedback. %; it is domain-agnostic and applies to both open-source and closed-source models. 
%\item The method is extended to self-refinement in multi-turn synthetic dialogue generation through both query and response generation. 
%Fine-tuning of strong instruction-tuned \textsc{llama-2-13b-chat} on the synthetic data further demonstrates the effectiveness of the algorithm.
%and fine-tune strong instruction tuned \textsc{llama-2-13b-chat} on this data to improve model's understanding of the principles and refinement through the difference in response quality. 

\item \textbf{ProMiSe} is applied to both content-grounded single-turn question answering and multi-turn dialogue generation. Extensive evaluations on MultiDoc2Dial and QuAC datasets with 5 automatic evaluation metrics (RougeL, Bert-Recall, Bert-K-Precision, Recall, K-Precision) as well as LLM-as-judge with GPT-4, demonstrate its effectivenss both in few-shot and fine-tuning settings. We will release both the software and the synthetic dialogue data.

%\item We apply ProMiSe for content-grounded question answering and demonstrate its efficacy both with and without additional training (i.e. few-shot) on the MultiDoc2Dial and QuAC datasets, and with GPT-4 as an evaluator.
\item We analyze the relationship between the change in proxy metric scores and the downstream evaluation metrics, revealing an unsupervised correlation and reinforcing the efficacy of our method.
%\item We perform ablation studies to examine the relationship between metric threshold calibration as a stopping criterion and the response over-optimization phenomenon.
\end{itemize}

\begin{algorithm}[tb]
\caption{\textit{ProMiSe Self-Refinement}}\
\footnotesize
\begin{algorithmic}
    \STATE {\textbf{Inputs:}} Model $\mathcal{M}$;
    \STATE \hspace{2mm} $x$: Task Input (document, context, $\dots$);
    \STATE \hspace{2mm} $\mathcal{P}$: User-defined set of principles;
    \STATE \hspace{2mm} $\mathcal{T}$: Set of metrics corresponding to $\mathcal{P}$;
    \STATE \hspace{2mm} $i$: Initial generation prompt;
    \STATE \hspace{2mm} $r_p$: Refinement prompt for principle $p$;
    \STATE \hspace{2mm} $\tau = [\tau_1, \tau_2, \dots, \tau_{|\mathcal{T}|}]$: Metric $i \in \mathcal{T}$ has threshold $\tau_i$;
    \STATE \hspace{2mm} $w = [w_1, w_2, \dots, w_{|\mathcal{T}|}]$: Metric $i \in \mathcal{T}$ has weight $w_i$;
    \STATE \hspace{2mm} $\lambda$: Improvement threshold for weighted metric sum;
    \STATE \hspace{2mm} $N$: Number of initial responses generated per turn
    \STATE
    \STATE {\textbf{Begin:}}
\end{algorithmic}
\begin{algorithmic}[1]
    \STATE $\mathcal{Y}_0 = \{y_n\}_{n=1}^N$ with $y_n \sim p_{\mathcal{M}}(y \mid x, i)$
    \STATE $y^0 = \argmax\limits_{y \in \mathcal{Y}_0}\left\{\sum\limits_{t=1}^{|\mathcal{T}|} \mathbbm{1}(\argmax\limits_{y' \in \mathcal{Y}_0}\{m_t(y', x)\} = y) \right\}$
    \IF{$\sum\limits_{t=1}^{|\mathcal{T}|} \mathbbm{1}(m_t(y^0, x)\geq \tau_t) = |\mathcal{T}|$}
        \STATE \textbf{return} $y^0$
    \ELSE
        \FOR{iteration $j = 0, 1, \dots J$:}
            \STATE $\mathcal{Y}_{j+1} = \{y_p\}_{p=1}^{|\mathcal{P}|}$ with $y_p \sim p_{\mathcal{M}}(y \mid y^j, x, r_p)$
            \STATE \footnotescript{$y^{j+1} = \argmax\limits_{y \in \mathcal{Y}_{j+1}}\left\{\sum\limits_{t=1}^{|\mathcal{T}|} \mathbbm{1}(\argmax\limits_{y' \in \mathcal{Y}_{j+1}}\{m_t(y', x)\} = y) \right\}$}
            \footnotesize
            \IF{$\sum\limits_{t=1}^{|\mathcal{T}|} \mathbbm{1}(m_t(y^{j+1}, x)\geq \tau_p) = |\mathcal{T}|$}
                \STATE \textbf{return} $y^{j+1}$
            \ELSIF{$\sum\limits_{t=1}^{|\mathcal{T}|} w_t \cdot \mathbbm{1}(m_t(y^{j+1}, x) \geq m_t(y^{j}, x))$} %< \lambda$}
                \STATE $y^{j+1} = y^j$
            \ENDIF
        \ENDFOR
        \STATE \textbf{return} $y^J$
    \ENDIF
\end{algorithmic}
\end{algorithm}
\vspace{-2mm}
\section{Algorithm}
%At a high level: given an input (e.g. conversation history), our algorithm identifies a promising candidate response via rejection sampling, obtains external feedback through proxy metrics, refines the response with respect to each principle if deemed insufficient and selects the most promising refinement, and proceeds with the refinement iteratively until termination.
%At a high level, 
Given an input, e.g. a document and conversation history, ProMiSe executes three main steps: (i) Generate an initial response, (ii) Obtain external feedback via proxy metrics, and (iii) Refine the response with respect to each principle, if the response is deemed inadequate by the feedback mechanism. The last two steps are run iteratively until the response meets a quality threshold. We present a detailed description of these steps below. 

\subsection{Initial Response Generation} 
\label{subsec:inital-response}
For an input instance, we perform Best-of-N sampling to yield a set of %\footnote{We find that Best-of-N Sampling is effective in achieving a higher quality initial response, via holdout set analysis. The objective in such an approach is based on the same proxy feedback mechanism as determining refinement.} 
 responses, $\mathcal{Y}_0$, from Language Model $\mathcal{M}$, given the input and an initial generation prompt. The initial generation prompt consists of an instruction and optional in-context demonstrations.
 %of response generation for the task. 
 The instruction explicitly suggests that a response be generated which reflects desirable principles, the set of which is contained in $\mathcal{P}$. We determine the quality of the sampled responses based on a set of proxy metrics determined a priori, designated as $\mathcal{T}$. %, where $|\mathcal{T}| = k$. 
 We note that the selected metrics should be designed by the user to improve alignment by reflecting the principle set for the response, $\mathcal{P}$, with respect to the current task. As such, each metric $m_t$ is also predicated on the inputs provided which may be used as means to assess candidate responses -- a text passage or document, conversation history, etc. (thus lending itself to the content-grounded setting). Each responses in $\mathcal{Y}_0$ is scored with each metric $m_t$ in $\mathcal{T}$, and the response with the highest scores on the greatest number of metrics is chosen as the best initial response, $y^0$.
 %Each response $R$ is scored with each metric $i$ in $\mathcal{T}$, yielding a score vector $S_{R} = [S_{R, i}]_{i \in \mathcal{T}}$. The response with the highest scores on the majority of the metrics is chosen as the best initial response, denoted $R_t$. 
 
Next, we determine the global sufficiency of $y^0$ as an acceptable response, by comparing the proxy scores element-wise against a threshold $\tau$, consisting of scalar values $\tau_1, \tau_2, \dots, \tau_{|\mathcal{T}|}$. $\tau_i$ is the minimum value such that a response is deemed sufficient, for each metric $i$ in $\mathcal{T}$. If the scores of $y^0$ exceeds their respective thresholds, for all $\mathcal{T}$ components, we return it as the final response. If not (i.e. $y^0$ fails to clear the threshold on at least one metric), we proceed to the refinement module. 

%If $S_{R_t} \geq \tau$ (that is, $S_{R,t} \geq \tau_i$ for all metrics $i \in \mathcal{T}$), then we take $R_t$ as the final response and proceed to instance $t+1$; if not, we continue to the refinement module. 

%\begin{figure}[h]
%    \centering
%    \includegraphics[scale=0.35]{self-refinement-algo-revised.png}
    %\includegraphics[scale=0.4]{self-refinement-algo.png}
    %\caption{}
%    \label{fig:enter-label}
%\end{figure}
%\vspace{-6mm}
%\vspace{-2mm}

\subsection{Response Refinement} 
\label{subsec:response-refinement}
Our approach to response refinement is predicated on in-context exemplars of principle-specific refinement for the given task. The refinement prompt also contains the previous best response, denoted $y^j$ — in the first iteration, this is equal to $y^0$.
%which was deemed insufficient based on the external proxy metrics.
Aligning responses with multiple principles (i.e. where $\mathcal{P} > 1$) induces a multi-objective problem; rather than explicitly optimizing across the set simultaneously, we propose deliberate refinement with respect to one principle at a time, selecting an optimal candidate at each iteration based on the proxy metrics. For each iteration in the self-refinement phase, we loop through the set of principles $\mathcal{P}$ and generate a set of new responses, with the goal of each resulting response reflecting improvement on its respective principle. In each such query to Language Model $\mathcal{M}$, we introduce a principle-specific refinement prompt, consisting of in-context demonstrations of refinement and an instruction to improve the current best response, both with respect to the current principle. Examples of such prompts are contained in Appendix \ref{app:section-C}. %Appendix C.

%\paragraph{Optimal Refinement Selection via Majority Voting} Given a set of refinement candidates, we aim to select a single response which would be most likely to be deemed sufficient relative to the threshold $\tau$ by improving across the set of principle objectives. To this effect, considering the simplex of principle dimensions, we posit that the “centroid” among these candidates is the most well-rounded. We apply ensemble consensus majority voting with a symmetric distance metric; note that this metric does not necessarily have to be contained in $\mathcal{T}$, but should be reflective of the principles in $\mathcal{P}$. This approach also helps to mitigate instances where there is simultaneous improvement along one dimension (e.g. relevance) but worse quality with respect to another (e.g. faithfulness). The returned response, chosen as the best refinement, is denoted $R_{t,j}$, for iteration $j$ of the refinement procedure. 

\paragraph{Determining Improvement} 

We perform rejection sampling, this time on the set of refinement candidates, scoring with each metric in $\mathcal{T}$ and selecting the response, $y^{j+1}$, with the highest scores on the majority of metrics. The scores of $y^{j+1}$, the best refinement candidate, are then compared against the threshold $\tau$. If the scores of $y^{j+1}$ exceed the threshold on all $|\mathcal{T}|$ metrics, then we stop refinement and accept it as the final response. Otherwise, we now compare against the scores of the previous best response, $y^j$. The user assigns weights $w = [w_1, w_2, \dots, w_{|\mathcal{T}|}]$ for the respective metrics in $\mathcal{T}$; these importances should likely be informed by the principles in $\mathcal{P}$ which each metric corresponds to, and the user's design goals. Then, given the scores for $y^{j+1}$ and $y^j$, we compute: $$\sum_{t=1}^{|\mathcal{T}|} w_t \cdot \mathbbm{1}(m_t(y^{j+1}, x) \geq m_t(y^{j}, x))$$ For each metric in $\mathcal{T}$, the indicator takes on a value of 1 if the new response is an improvement on the previous best response, with respect to that metric, or 0 otherwise, and is weighted by the elements in $w$. If this sum fails to exceed a user-defined threshold of $\lambda$, we
do not update the best refinement response for this iteration (i.e. set $y^{j+1} = y^j$); else, we proceed to the next refinement iteration, until termination. 

\section{Evidence: Question Answering}
\label{sec:evidence}

%An area of direct application for our approach is 
We apply ProMiSe to content-grounded question answering: given a document and a conversation history, which may consist of a single user utterance (a question posed to the conversational agent) or a multi-turn dialogue between the user and the agent, we seek for the LLM to produce a response to the most recent user query.

%We selected content-grounded question answering as a ripe application task for our algorithm. The objective is: given a document and a conversation history, which may consist of a single user utterance (a question posed to the conversational agent) or a multi-turn dialogue between the user and the agent, we seek for the Large Language Model to produce a response to the most recent user query using the document. In other words, \textit{a factual response should be grounded to the provided document}. As this category may be explored through either Extractive QA or Abstractive QA, we evaluate on datasets consisting of gold responses reflective of both settings. 

\subsection{Set of Principles} 
\label{subsec:priniciples}
We first identify an appropriate set of principles for the task, which define key characteristics of a good agent response. They are as follows:
%For this task, we identify a set of crucial, distinguishable principles that constitute a response of sufficient quality (there is no implicit ordering of importance among the principles, reflected in the refinement loop of Algorithm 1): 

\begin{enumerate}
\item \textbf{Specificity.} 
%As vague answers are generally not preferable, we would like for the agent’s response to capture nuances and details to yield a more concrete answer. 
If an initial response is too vague, this would likely lead to more user interactions asking the agent to make its response more specific. 
%which we seek to reflect with our refinement prompt. 
\item \textbf{Faithfulness.} We suggest that accurate, factual responses are those grounded in the document, and thus should have high (semantic and lexical) overlap with the document. 
\item \textbf{Relevance and Consistency.} The conversational agent response should be relevant to the most recent user query, and by induction to the entire conversation history.

%The conversational agent response should be relevant to the most recent user query and the conversation history as a whole; the response should be targeted to the question that was posed. It is to be noted that consistency with the conversation history is implicitly captured through this in an inductive manner: if all previous utterances (including the query) were consistent, then relevance to the query ensures consistency of the full dialogue. 
\end{enumerate}

\subsection{In-Context Demonstration Selection}
\label{subsec:incontext-demo}

We explore our algorithm through the generation of both a single agent response and an entire multi-turn dialogue. 
%The latter requires generating both user and agent utterances through in-context learning; the full algorithm is included in Appendix A. %
We include the algorithm for multi-turn dialog generation in Appendix \ref{app:section-A}. %Appendix A.

\paragraph{Response and Query Generation.} For the generation of an initial response consistent with the content-grounded QA setting, we extract 3 instances from the MultiDoc2Dial \cite{multidoc2dial} training data as in-context exemplars; the prompt template is included in Appendix \ref{app:section-C}. This includes the document, conversation history, and the gold response provided by the annotators. The in-context exemplars for query generation work similarly, with 3 demonstrations consisting of different conversation lengths (in number of utterances), but where the last utterance is the final user query. 
%With recent findings suggesting that in-context learning relies on structure of the demonstrations more than the specific content, we leverage this to enable domain-agnostic exemplar selection for our application, generalizing across datasets and supported by our experimental findings in Section 4. 

\paragraph{Principle Refinement.} To perform in-context refinement on a particular principle, we similarly take 3 in-context demonstrations from the training dataset, but seek to contrast between a better and worse response, with respect to the principle. 
%we seek to contrast the \textit{refinement delta} (that is, show the improvement between the worse response and better response with respect to the principle). 
To accomplish this, we manually annotate a worse response for each instance relative to the gold response. In the prompt, we model this as 3 separate utterances: the worse agent response, a user turn probing the agent to improve its response to update along the principle, and another agent utterance containing the better response (i.e. the gold response). To more explicitly suggest the presence of a response quality difference, we include the tags “not \{principle\}” and “more \{principle\}”, for the two agent turns, respectively, where \{principle\} is either ‘specific’, ‘relevant’, or ‘accurate’. 
%Our experiments with development sets also suggest that using the word “not” as opposed to “less” for the former tag better conveys the difference between a good and a bad response to the model.
%with regards to providing a tag for the worse response better conveys the difference between a good and a bad response to the model. 

%\input{inference_table}
\begin{table*}[t]
\vskip 0.15in
\begin{center}
%\begin{scriptsize}
%\begin{footnotesize}
\begin{footnotescript}
\begin{sc}
\begin{tabular}{lccccccr}
\toprule
 Proxy Metrics & Stage & Rouge-L & BERT-Recall & BERT K-Prec. & Recall & K-Prec. \\
\midrule
\textbf{MD2D} Flan-T5-XXL (11B) & & & & & & \\
\midrule
 Only Rou-L + Rou-1 & Initial & 21.55 & 28.11 & 40.42 & 32.34 & 76.77 \\
& Final & 21.72 & 29.29 & 42.74 & 34.14 & 79.29 \\
\hline
  Only RM & Initial & 22.33 & 28.91 & 44.58 & 33.61 & 81.29\\
 & Final & \textbf{22.43} & 29.17 & 45.60 & 34.20 & 82.25 \\
 \hline
 0-Shot / Rou + RM   & Initial & 22.30 & 28.94 & 44.55 &  33.68 & 81.56    \\
& Final & 22.38 & \textbf{30.10} & \textbf{46.60} & \textbf{35.58} & \textbf{83.13}\\
\hline\midrule
\textbf{MD2D} Llama-2-13B-Chat & & & & & & \\
\midrule
 Only Rou-L + Rou-1 & Initial & 19.31 & 28.92 & 34.44 & 38.45 & 70.33 \\
 & Final & 18.95 & 29.67 & 36.04 & 40.43 & 71.76        \\
\hline
 Only RM & Initial & 19.97 & 29.65 & 34.33 & 38.07 & 70.08 \\
& Final & 19.95 & 29.89 & 40.68 & 38.59 & 76.73 \\
\hline
 Rou + RM & Initial & \textbf{20.36} & 30.17 & 40.68 & 38.59 & 76.84 \\
& Final & 20.06 & \textbf{30.64} & \textbf{41.46} & \textbf{40.00} & \textbf{77.43} \\
\bottomrule
\midrule
\textbf{QuAC} Flan-T5-XXL (11B) & & & & & & \\
\midrule
 Only Rou-L + Rou-1 & Initial & \textbf{41.57} & 40.70 & 43.31 & 44.87 & 87.57 \\
& Final & 40.00 & \textbf{41.47} & 46.06 & \textbf{45.77} & 88.26 \\
\hline
  Only RM & Initial & 37.01 & 36.58 & 48.82 & 40.21 & 91.51\\
 & Final & 34.99 & 35.00 & 49.69 & 38.61 & 91.58 \\
 \hline
 Rou + RM   & Initial & 38.20 & 37.28 & 48.44 &  40.78 & 91.38    \\
& Final & 35.85 & 37.13 & \textbf{50.91} & 41.21 & \textbf{91.89} \\
\hline\midrule
\textbf{QuAC} Llama-2-13B-Chat & & & & & & \\
\midrule
 Only Rou-L + Rou-1 & Initial & \textbf{31.36} & 35.12 & 40.79 & 42.86 & 83.08 \\
 & Final & 29.23 & \textbf{35.28} & 42.87 & \textbf{43.63} & 82.96       \\
\hline
 Only RM & Initial & 29.83 & 33.11 & 46.64 & 39.22 & 87.78 \\
& Final & 28.70 & 31.83 & 47.79 & 37.62 & 87.24 \\
\hline
 Rou + RM & Initial & 28.85 & 32.29 & 46.59 & 38.54 & 88.04 \\
& Final & 26.76 & 32.39 & \textbf{48.05} & 39.64 & \textbf{88.11} \\
\bottomrule
\end{tabular}
\end{sc}
%\end{scriptsize}    
%\end{footnotesize}
\end{footnotescript}
\end{center}
\vskip -0.1in
\caption{Experimental Results on the MultiDoc2Dial (MD2D) and QuAC test sets, containing 10,204 and 1,000 instances, respectively. Experiments are reported with the Flan-T5-XXL (11B) and Llama-2-13B-Chat models, using 3 Rouge (ROU) measures, the WeCheck reward model (RM), and both in tandem for thresholding. "Initial" refers to scoring generations after rejection sampling, while "Final" includes both "sufficient" initial responses and post-refinement responses. In proxy metrics, Rouge-L includes computing between the candidate response with both the grounding document and the given user query, and Rouge-1 is with respect to the document. Highest scores are boldfaced for each model. We decode with sampling method by setting temperature=0.7, top-k=50 and top-p=1. 
}
\label{tab:combined_inference}
\end{table*}

\subsection{External Proxy Metrics} 
\label{subsec:proxy-metric}

To capture the aforementioned principles, we define relevant proxy metrics. 
%that are cheap and efficient to compute. %This allows us to integrate a fundamental linguistic motivation associated with each metric. 
%It is crucial to select these proxies in a way that does not directly optimize the final evaluation metrics for our datasets, yet is still reflective of response quality improvement specifically along our chosen dimensions (principles). Simultaneously, we should ensure that we are not overfitting to improvement on proxy metrics (that is, the response improves in score via refinement on the intermediate metrics, but becomes worse on the final evaluation metrics), and thus use a validation set to test various combinations of the proposed metrics and determine sufficiency threshold values. 
The proxy metrics should be reflective of response quality improvement along our chosen dimensions and should not directly optimize the final evaluation metrics.
\vspace{-1.5mm}
\paragraph{ROUGE Metrics.} We select three ROUGE metrics intended to correspond to each of the three principles. ROUGE-1 recall between the response and the document mostly represents specificity as more specific answers contain more details from the document. Next, we use ROUGE-L between the response and the document — this primarily addresses faithfulness, as a greater score would suggest a more extractive answer, which is clearly preferable to hallucinated facts. Finally, we compute ROUGE-L between the response and the conversation history to capture consistency between the user query and the response and relevance of the response to the query history.
%We note that the Rouge-L measure presented in the evaluation results on MultiDoc2Dial is performed between the candidate response and the gold annotated response, and thus differs from our intermediate signals.

\begin{table*}[t]
\vskip 0.15in
\hspace{-50mm}
\begin{center}
%\begin{scriptsize}
\begin{footnotesize}
\begin{sc}
\begin{tabular}{lcccccc}
\toprule
Fine-tuning Data & SynsetSize & Rouge-L & BERT-Recall & Recall & BERT-K-Prec. & K-Prec. \\
\toprule
None (Baseline) & 0  & 21.11 & 30.95 & 38.62 & 40.05 & 76.89 \\
\hline
$Synset_{1}$-initial & 8k & 24.63 & 28.51 & 34.21 & 41.91 & 78.26  \\
$Synset_{1}$-final & 8k & 26.13 & 33.65 & 39.49 & 46.98 & 82.74 \\\hline
$Synset_{2}$-initial & 10k  & 24.51 & 27.82 & 33.81 & 41.01 & 76.71 \\
$Synset_{2}$-final & 10k & 26.06 & \textbf{33.83} & \textbf{40.56} & \textbf{48.67} & \textbf{84.46} \\\hline
$Synset_{3}$-initial & 14k & 26.18 & 30.43 & 34.81 & 41.73 & 79.08 \\
$Synset_{3}$-final & 14k & \textbf{26.78} & 33.57 & \textit{38.26} & 46.32 &      83.68  \\
\hline\hline
Human (Baseline) &  & 55.32 & 56.43 & 56.75 & 31.38 & 75.22 \\
\hline
Human+$Synset_{1}$-final & 8k & 55.40 & \textbf{57.24} & \textbf{57.79} & 32.53 & \textbf{76.13} \\
Human+$Synset_{2}$-final & 10k & \textbf{55.58} & 57.18 & 57.63 & 31.79 & 75.46  \\
Human+$Synset_{3}$-final & 14k & 55.00  & 56.92 & 57.56  & \textbf{32.78} & 75.87   \\
\bottomrule
\end{tabular}
\end{sc}
\end{footnotesize}
%\end{scriptsize}
\end{center}
\vskip -0.1in
\caption{Effectiveness of the proposed refinement algorithm measured by the synthetic data qualities. We QLoRA fine-tune \textsc{Llama-2-13B-Chat} model on the two sets of synthetic multi-turn dialogues, one generated with the refinement algorithm denoted by $Synset_{x}$-\textsc{final}, and the other generated without the refinement algorithm denoted by $Synset_{x}$-\textsc{initial}. $Synset_{1}$ includes 8k and $Synset_{2}$, 10k samples of 2 turn dialogues. $Synset_{3}$ includes 10k samples of 2 turn, 2k samples of 4 turn, and 2k samples of 6 turn dialogues.
The upper portion of the table compares the performance of the model fine-tuned on the synthetic data with the highest-scoring baseline without fine-tuning.  The lower portion of the table compares the performance of the model fine-tuned on the combination of human annotated and synthetic data with the model fine-tuned on human annotated data only.}
\label{tab:sdgtable}
\end{table*}

\vspace{-1mm}
\paragraph{WeCheck: Factual Consistency Checker.} Given a candidate response and the grounding document, WeCheck \cite{wecheck} addresses the faithfulness principle. 
%We make use of an external reward model, WeCheck \cite{wecheck}, as a factual consistency checker, given a candidate response and the grounding document — this addresses the faithfulness principle. 

Our experiments evaluate each model in three thresholding settings: solely using the three aforementioned ROUGE metrics, solely using the WeCheck model, and using a combination of both. If we use only WeCheck, rejection sampling is performed to yield the highest scoring response according to WeCheck
%sufficiency is only with respect to a threshold on WeCheck score, 
and we determine whether a refined response constitutes an improvement solely using the WeCheck scores. If using both ROUGE and WeCheck, a sufficient response must clear the threshold on all four metrics. During refinement, we yield a reward indicator with each category (Rouge and reward model, i.e. WeCheck) which is 1 if deemed to have improved during the present iteration and 0 otherwise, and compute a weighted sum using a user-defined weight vector $w$. If this sum is greater than 0.5, we update the best response to be the new one, else retain the previous best. 

%\input{sdgtable}
\begin{comment}
\subsection{Evaluation Setup}

 To evaluate our algorithm for content-grounded question answering, we use two widely-used open-source language models, Flan-T5-XXL \cite{flan-t5-palm} and Llama-2-13B-chat \cite{llama2} with 11B and 13B parameters, respectively.

\paragraph{Evaluation Datasets.} We evaluate on the MultiDoc2Dial \cite{multidoc2dial} and QuAC \cite{quac} datasets. %, particularly chosen to capture both Abstractive and Extractive QA through their gold responses. 
Both datasets feature conversations %and domain-specific questions 
wherein answers to queries posed by the user are expected to come from a document. %— this particularly highlights the domain-agnostic nature of our few-shot approach. 
We use a sub-document split on MultiDoc2Dial, to remove the information retrieval (IR) component such that we only have the most relevant document as opposed to the entire set of candidate documents; this also enables inclusion of multiple in-context exemplars under a limited context length. 

\paragraph{Multi-turn Dialogue.} %We furthermore generate 8k, 10k, and 14k instances of synthetic multi-turn dialogues with Flan-T5-XXL.
We generate synthetic dialogues of varying lengths from Flan-T5-XXL, containing refinement instances: the initial response, a user query to improve the response along a principle, and the refined response. We fine-tune Llama-2-13B-Chat on this data to measure the quality of synthetic data yielded from the proposed algorithm.
%effectiveness of the proposed algorithm on synthetic data qualities.
\end{comment}

\section{Experimental Results and Discussion}

We use two widely-used open-source language models to evaluate our algorithm for content-grounded question answering, \textsc{Flan-T5-XXL} \cite{flan-t5-palm} and \textsc{Llama-2-13B-chat} \cite{llama2} %(Apache 2.0 and Llama License).

\paragraph{Evaluation Datasets.} We evaluate the technique on the test dataset of MultiDoc2Dial \cite{multidoc2dial} (https://doc2dial.github.io/multidoc2dial/), content-grounded dialogues, and the validation dataset of QuAC \cite{quac} (https://quac.ai/), short form question-answering. %, particularly chosen to capture both Abstractive and Extractive QA through their gold responses. 
Both datasets feature conversations %and domain-specific questions 
wherein answers to queries posed by the user are expected to come from a document.\footnote{We use a sub-document split on MultiDoc2Dial, to remove the information retrieval (IR) component such that we only have the most relevant document as opposed to the entire set of candidate documents. We use the validation dataset of QuAC as the test data since the testset is not publicly available.} 

\paragraph{Evaluation Metrics}

We use five automatic evaluation metrics: ROUGE-L \cite{rouge2004}, BERTScore
Recall, BERTScore K-Precision (K-Prec. hereafter), \cite{bertscore2020iclr}, Recall, and K-Precision. ROUGE-L, BERTScore Recall and Recall measure the agreement between the candidate response and the provided gold response. BERTScore K-Prec. and K-Prec. measure the agreement between the candidate response and the grounding document. We chose Recall and K-Prec. metrics due to their strong correlation with human assessments of instruction-following models in content-grounded QA tasks, \cite{adlakha2023}.

\subsection{Single-Turn QA Results}
Table~\ref{tab:combined_inference} presents the results across the three possible metric sets (ROUGE metrics, the WeCheck reward model, and both) as defined in Section~\ref{subsec:proxy-metric}. It can be observed that designating $\mathcal{T}$ to be the combination of the three ROUGE metrics and the WeCheck reward model %, when defined as the set $\mathcal{T}$, 
yields improved results across the majority of the metrics. We find that using the WeCheck reward model as the sole sufficiency metric yields less consistent improvement across the set of evaluation metrics, yet boosts performance when applied in tandem with ROUGE metrics.

To identify the appropriate sufficiency threshold for the proxy metrics, we perform a rigorous study of various settings, included in Appendix \ref{app:section-B}. Experiments containing ROUGE metrics maintain a sufficiency threshold of 0.02 for response-document Rouge-1 Recall, 0.05 for response-document Rouge-L and 0.05 for response-query Rouge-L. Results involving the WeCheck reward model use a threshold of 0.5 between the response and document. 

%It was observed that specificity refinement along with the chosen metrics was, in fact, sufficient to capture all three desirable principles — that is, refinement on “specific” yielded more faithful (to the document) and relevant responses in the process. At the same time, refinement on faithfulness yields minimal gains in accuracy, often reproducing the same response. This appears to corroborate with recent findings regarding overconfidence and honesty \cite{yang2023alignment, zhang2023rtuning}; that is, if the LLM believes that the response is already sufficient (when it is not, as inadequacy of the response is necessary to proceed to the refinement phase in our algorithm), it will not improve said response. 

\begin{table}[h]
\label{tab:avg-lengths}
%\vskip 0.15in
%\hspace{-50mm}
\begin{center}
\begin{footnotesize}
%\begin{scriptsize}
%\begin{sc}
\begin{tabular}{lccr}
\toprule
 & Initial &  Final \\
 \midrule
 \textbf{MultiDoc2Dial} (Avg. Gold: 15.55) & & \\
\midrule
Flan-T5-XXL ROU-Only & 32.06 & 35.88 \\
\midrule 
Flan-T5-XXL RM-Only & 33.16 & 33.70 \\
\midrule
Flan-T5-XXL ROU + RM & 33.00 & 36.09 \\
\midrule
Llama-2-13B-Chat ROU-Only & 39.30 & 44.40 \\
\midrule 
Llama-2-13B-Chat RM-Only & 38.45 & 39.51 \\
\midrule
Llama-2-13B-Chat ROUGE + RM & 38.56 & 42.59 \\
\midrule\midrule
\textbf{QuAC} (Avg. Gold: 12.57) & & \\ 
\midrule
Flan-T5-XXL ROU-Only & 17.40 & 20.46 \\
\midrule 
Flan-T5-XXL RM-Only & 18.19 & 19.07 \\
\midrule
Flan-T5-XXL ROU + RM & 17.79 & 21.49 \\
\midrule
Llama-2-13B-Chat ROU-Only & 29.61 & 33.41 \\
\midrule 
Llama-2-13B-Chat RM-Only & 27.73 & 27.39 \\
\midrule
Llama-2-13B-Chat ROU + RM & 28.74 & 32.58 \\
\bottomrule
\end{tabular}
%\end{sc}
%\end{scriptsize}
\end{footnotesize}
\end{center}
\vskip -0.1in
\caption{Average word token counts for initial and final generations with ProMiSe. Statistics are computed for the three different settings of the proxy metric set, $\mathcal{T}$; RM is the WeCheck reward model.}
\label{tab:wordcount}
\vspace{-3mm}
\end{table}

%In Table 1, we also compare based on the number of few-shot exemplars employed at the initial response generation phase. We found that the 0-shot performance of initial responses is, in fact, higher than 3-shot results, suggesting that instruction-tuned LMs such as Flan-T5-XXL are already fairly adept at dialogue response generation. Simultaneously, we found that 3-shot results with refinement constitute an improvement over 0-shot performance; that is, there is still room for improvement after fairly coherent initial responses, which is achieved when using 3 exemplars per principle. 
%Thus, the results reported hold 3 exemplars constant for the refinement phase. 

In Table~\ref{tab:wordcount}, we compare the average length (word count) of the initial and final responses, for the MultiDoc2Dial and QuAC datasets. It can be observed that the length of final responses is marginally greater than the average initial response length. This suggests that our performance improvements exhibited in Table 1 are unlikely to be solely a result of longer responses (e.g. reproducing large sections of the document). Simultaneously, our model producing longer responses relative to the gold response likely explains slight declines in Rouge-L scores with both models and both datasets; in particular, Llama-2's responses are much longer than Flan-T5's and the gold response.

\paragraph{Analysis with Proxy Metrics}

%In addition to the observation that using the WeCheck reward model as the sole sufficiency metric yields less consistent improvement across the set of evaluation metrics, yet boosts performance when applied in tandem with ROUGE metrics, 

We explore the relationship between the improvement in the final evaluation metrics and the direction of change on the proxy metrics in ProMiSe. That is, is improvement on the proxy sufficiency metrics during the execution of the algorithm correlated with the downstream evaluation metric improvement from initial to final response? %as reflected by the downstream evaluation metrics?  %Simultaneously, we wish to determine: which proxy metric(s) are most influential in causing refinement (and in turn, which principles are most lacking in the initial responses generated)?

\begin{table}[h]
%\vskip 0.15in
%\hspace{-50mm}
\begin{center}
%\begin{scriptsize}
%\begin{footnotesize}
\begin{footnotescript}
%\begin{sc}
\begin{tabular}{lccccr}
\toprule
 & Count & ROU-L Diff. & BERT-R Diff. \\
 \midrule
 \textbf{ROU-Only} & & &  \\
\midrule
ROU 3 $\uparrow$ & 376 & +2.31\% & +8.26\% \\
\midrule
ROU 2 $\uparrow$ & 128 & +2.41\% & +6.81\% \\
\midrule
ROU 1 $\uparrow$ & 69 & +0.64\% & -1.74\% \\
\midrule
\midrule
\textbf{RM-only} & & & \\
\midrule
RM $\uparrow$ & 169 & +0.68\% & +1.68\% \\
%\midrule
%RM-only, $\leftrightarrow$ & 1869 & +0\% & +0\% \\
\midrule
\midrule
\textbf{ROU + RM} & & & \\
\midrule
ROU 3 $\uparrow$, RM $\uparrow$ & 110 & +2.21\% & +8.23\% \\
\midrule
ROU 2 $\uparrow$, RM $\uparrow$ & 45 & +0.25\% & +1.57\% \\
\midrule
ROU 1 $\uparrow$, RM $\uparrow$ & 59 & +1.18\% & -0.66\% \\
\midrule
ROU 3 $\uparrow$, RM $\downarrow$ & 195 & -1.40\% & +5.67\% \\
\midrule
ROU 2 $\uparrow$, RM $\downarrow$ & 50 & +0.07\% & +4.25\% \\
\midrule
ROU 1 $\uparrow$, RM $\downarrow$ & 32 & +3.37\% & +2.16\% \\
% ROUGE-only, 3 $\downarrow$ & 0 & +0\% & +0\% \\
% \midrule
%ROUGE+RM, All 3 $\uparrow$ & 347 & -0.13\% & +6.47\% \\
%\midrule
%ROUGE+RM, 2 $\uparrow$ & 97 & -0.14\% & +2.99\% \\
%\midrule
%ROUGE + RM, 1 $\uparrow$ & 127 & +1.45\% & +0.33\% \\
%\midrule
%ROUGE+RM, All 3 $\downarrow$ & 7 & -1.91\% & -11.00\% \\
\bottomrule
\end{tabular}
%\end{sc}
%\end{footnotesize}
%\end{scriptsize}
\end{footnotescript}
\end{center}
\vskip -0.1in
\caption{Analysis of the correlation between improvement on proxy ROUGE (ROU) and WeCheck reward model (RM) metrics with change in the final evaluation ROUGE-L and BERT-Recall with the gold response. Performed with Flan-T5-XXL on a 2,038 sample MultiDoc2Dial development set. Proxy metric scores are computed between the candidate and either the provided document or context. $\uparrow$ and $\downarrow$ represents improvement and decline, respectively. "2 $\uparrow$" means that two of the proxy ROUGE metric set improved. The differences reported are averaged across the sample count.}
\label{tab:unsupervised}
\vspace{-2mm}
\end{table}

The relationship between the proxy metric scores and the Rouge-L and BERTScore-Recall evaluation metrics is shown in Table~\ref{tab:unsupervised}. 
%The former analyzes the three proxy ROUGE metrics; that is, when said set is included in $\mathcal{T}$, we examine the average difference for the change in the two evaluation metrics. Table 4 studies the change in the WeCheck reward model score and its correlation with the evaluation metrics. 
We find that the chosen proxy metrics appear to serve as an unsupervised link %from our determining sufficiency as a means of external feedback 
to the final evaluation metrics. The number of samples that improve for each proxy metric change are roughly similar, a trend noticeable across settings. Furthermore, a greater degree of improvement on proxy metrics (e.g. improving on all three ROUGE metrics) generally corresponds to a larger average improvement (or less negative change) for Rouge-L and BERTScore-Recall with respect to the gold response. This highlights the value of our external metric feedback technique: by optimizing on a scoring scheme while simultaneously preserving the integrity of the downstream evaluation metrics, we can capture a similar notion of response quality and sufficiency.

\subsection{Multi-Turn Synthetic Dialogues}

%\paragraph{Multi-turn Dialogue.} %We furthermore generate 8k, 10k, and 14k instances of synthetic multi-turn dialogues with Flan-T5-XXL.
We generate synthetic dialogues of varying lengths from Flan-T5-XXL, containing refinement instances: the initial response, a user query to improve the response along a principle, and the refined response. 
The dialogues are generated from scratch, bootstrapping solely on the grounding documents in MultiDoc2Dial training data. 
%as a means to generate domain-relevant user queries and faithful agent responses according to the refinement algorithm using \textit{Rouge+RM} metrics. 
They alternate between user and agent utterances, and consist of 1-3 agent responses (thus containing total 2, 4, or 6 turns). We sampled 10k dialogues with 2 turns, 2k with 4 turns and another 2k with 6 turns. We QLoRA fine-tune \cite{dettmers2023qlora} Llama-2-13B-Chat model on these synthetic data. See Section~\ref{sec:fine-tuning} for details.

The results are shown in Table~\ref{tab:sdgtable}.
We observe sizable improvements across all metrics when comparing the performance without refinement, denoted \textsc{initial}, as opposed to with refinement, denoted \textsc{final}. Notably, these improvements are present on both lexical and semantic similarity measures; +6-6.75\% for both BERT-Recall and Recall, and +7.5-8\% for BERT K-Precision and K-Precision. Furthermore, merging the synthetic data with 38k samples of human annotated data from the MultiDoc2Dial train set  yields improvements over solely training on human annotated data. 
These results suggest the value of response quality refinement in generating high-quality synthetic data and yielding downstream improvements on evaluation metrics.

\begin{figure}[h]
    \centering
    \includegraphics[scale=0.40]{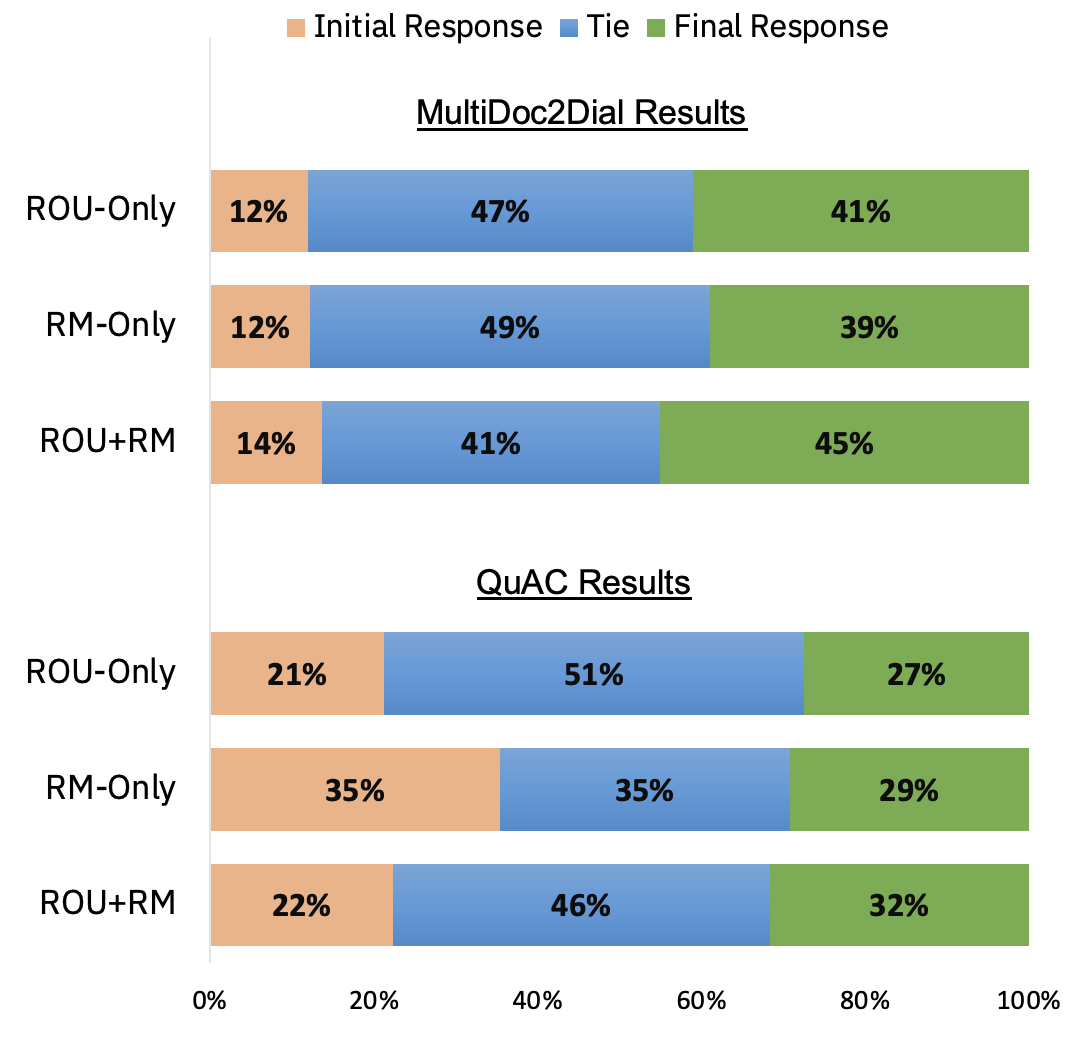}
    \caption{GPT-4-as-a-Judge results on Flan-T5-XXL for MultiDoc2Dial (MD2D) and QuAC. With 2551 randomly sampled instances from the MultiDoc2Dial test set, we examine those for which the initial and final response differ: 495 samples for ROUGE-only, 131 samples for RM-only (WeCheck), and 504 samples for ROUGE + RM. We perform a similar analysis with all 1000 QuAC test set instances; the respective counts are: 193 samples for ROUGE-only, 65 samples for RM-only, and 224 samples for ROUGE + RM. 
    }
    \label{fig:gpt-judge}
\vspace{-4mm}
\end{figure}

\subsection{LLM-as-a-Judge Evaluation}

We also perform automated evaluation with GPT-4 as a judge,  \cite{zheng2023judging}, which has been shown strongly correlate to human evaluation. Given the initial and final generations, we prompt the model to impartially assess which response is better. We largely adapt the prompts used for MT-bench evaluation in \citet{zheng2023judging}, which we show in Appendix \ref{app:section-F}.
%given the information in the document and the conversation history thus far. The model should output A (the first response is better), B (the second response is better), or C (the responses are tied in quality). The order in which the responses are presented is randomly permuted, and avoiding the influence of ordering on the decision-making of the LLM is explicitly modeled in the instruction; this can be seen in Appendix E. 
For MultiDoc2Dial, we randomly sample 2,551 of the indices of the test set responses (exactly one quarter), and only perform evaluation on samples for which the initial and final responses differ as a result of refinement. With the QuAC dataset, analyze all 1,000 test set instances, likewise evaluating where initial and final responses differ. The results are shown in Figure \ref{fig:gpt-judge}, where the numbers in percentage are the win rate of each response.

We find that GPT-4 deems the final response to be better than the initial response on all conditions for MultiDoc2Dial. The relative outlier is the QuAC dataset with RM-only; this is likely because WeCheck measures entailment rather than agreement.   Often the correct short response is less likely to be entailed than an incorrect longer response by the grounding document. However, a higher win rate with the ROUGE + RM combination validates the complementary nature of our proxy metrics. Furthermore, the strong correlation between the automatic evaluation metrics in Table~\ref{tab:combined_inference} with the GPT-4 evaluation results in Figure \ref{fig:gpt-judge} evidences the efficacy of our algorithm.

%We find that GPT-4 deems the final response to be at least as good as, if not better than, the initial generation on 86-88\% of samples for MultiDoc2Dial. 
%and is fairly consistent across the three proxy metric settings, as evidenced in Figure 2. 
%The strong correlation between the five automatic evaluation metrics in Table 1 with the GPT-4 evaluation results in Figure 2 serves further evidence of the efficacy of our algorithm -- self-refinement with ProMiSe often results in responses that are better or as good, rarely making them worse. A higher win rate with the ROUGE + RM combination further validates our proxy metrics.

\section{Other Related Work}
%There have been numerous work on self-refine, often dubbed as self-critique, self-improve, etc.
Various work on self-refinement may be distinguished according to the source of feedback, \cite{pan2023, huang2023large}. Internal feedback relies on the model's inherent knowledge and parameters to reassess its outputs. External feedback incorporates inputs from humans, other models. Our work is inspired by \cite{madaan2023selfrefine}. Unlike \citet{madaan2023selfrefine}, however, who rely on very large LLMs (GPT-3.5, ChatGPT, GPT-4) as the source of \textit{internal} feedback, we introduce \textit{external} feedback with proxy metrics and enable self-refinement technique to work with relatively small LLMs including Flan-T5-XXL and Llama-2-13B-Chat in content-grounded setups. 

Regarding internal feedback, \cite{constitutionalAI2022} experiment with method for training a harmless AI assistant through self-improvement. \cite{wang2023shepherd} propose Shepherd, a language model tuned to critique its own responses and suggest refinements. 
As for external feedback, \cite{paul2023refiner} propose \textsc{refiner}, a framework for finetuning LMs to explicitly generate intermediate reasoning steps while interacting with a critic model that provides automated feedback on the reasoning. \cite{gou2023tool} propose \textsc{critic} that interacts with appropriate tools, e.g. calculator, search engine, wikipedia, etc., to evaluate certain aspects of the text and then revise the output based on the feedback obtained during the validation process. %\cite{chen2023selfdebug} propose Self-Debugging, which teaches a large language model to debug its predicted program via few-shot demonstrations. 
\cite{olausson2024selfrepair} critically examines the LLM's ability to perform self-repair on problems taken from HumanEval and APPS and concludes that self-repair still lags behind what can be achieved with human-level debugging. \cite{gao2023acl} propose RARR (Retrofit Attribution using Research and Revision) that revises a generated text on the basis of the relevant evidence retrieved by re-search.

%\section{Related Work}
%\include{relatedwork}

\section{Conclusion}

We present a novel algorithm, ProMiSe, for self-refinement of language models. 
%Our algorithm has three key mechanisms: (i) Best-of-N rejection sampling to determine a best initial response, (ii) Proxy metric feedback, and (iii) Principle-specific refinement to incorporate feedback of insufficiency. 
%Unlike prior works which obtain \textit{intrinsic} feedback or \textit{external} single-aspect feedback to be incorporated into improving subsequent generations, 
ProMiSe uses \textit{external multi-aspect} feedback via proxy metrics capturing desirable principles for a high-quality response. 
%The unsupervised correlation between the selected external proxy metrics (with respect to a grounding document and context) and downstream evaluation demonstrates that greedily optimizing on individual principles at each refinement iteration corresponds with final response quality improvement. Fine-tuning on synthetic dialogues incorporating ProMiSe self-refinement reveals a significant increase in response quality, further evidencing its efficacy. 
ProMiSe is applied to content-grounded single-turn question answering and multi-turn dialogue generation. Extensive evaluations on MultiDoc2Dial and QuAC datasets with 5 automatic evaluation metrics as well as LLM-as-a-judge with GPT-4, demonstrate its effectiveness in both few-shot learning and supervised fine-tuning setups. This approach crucially enables relatively small LMs like Flan-T5-XXL and Llama-2-13B-Chat to successfully perform self-refinement. 
%Our analysis of the relationship between the proxy metric scores and the downstream evaluation metrics, reveals an unsupervised correlation, reinforcing the efficacy of our method.

%rather than relying on very large, closed-source models like GPT-3.5 and GPT-4. 
%We highlight the domain-agnostic nature of our few-shot approach: we retain the same provenance of our few-shot exemplars while using both the MultiDoc2Dial and QuAC datasets. Furthermore, the results of this work emphasize the value of deliberately selecting intermediate signals to reflect useful principles; as such, we suggest that ProMiSe may be extended to encompass more principles and be applied to other tasks.

\newpage
\section{Limitations}

Our work employs two open-source LMs: \textsc{flan-t5-xxl} and \textsc{llama-2-13b-chat}. Therefore, the generated data, including the synthetic multi-turn dialogues, can be susceptible to the limitations of such LMs, particularly the biases inherent in the training data which may be harmful with hate, abuse and social stereotypes. We have tested the algorithm ProMiSe on English only although it would have been more desirable to verify the value of the algorithm in multi-lingual setups. We have conducted extensive evaluations including 5 well-known automtic evaluation metrics and LLM-as-a-judge with GPT-4, which has been shown to correlate well with human evaluations. Nonetheless, inclusion of human evaluation would have strengthened our position further.

\section{Ethics and Impact}

Our technique can be used to guide generations towards user-specified targets; however, this could be applied to generate toxic or malicious content, by way of an adversarial principle selection. Nonetheless, we note that ProMiSe does present meaningful implications in enabling alignment to human preferences (where preferences, in this setting, refer to the user-defined principles). We will release the software for the ProMiSe algorithm, enabling others in the community to consider other principles of interest, or applications to other tasks. 

\bibliography{custom}

\appendix

\newpage
%\section{Example Appendix}
\label{sec:appendix}
\onecolumn

\section{Self-Refinement Algorithm for Synthetic Dialogue Generation} \label{app:section-A}

\begin{figure}[h]
\centering
\begin{minipage}{0.75\linewidth}
\begin{algorithm}[H]
\caption{Synthetic Dialogue Generation with ProMiSe}\
%\small
%\normalsize
\smallnormalsize
%\footnotesize
\begin{algorithmic}
    \STATE {\textbf{Inputs:}} Model $\mathcal{M}$;
    \STATE \hspace{2mm} $d$: Input document; 
    \STATE \hspace{2mm} $c = \emptyset$: conversation history
    %\STATE \hspace{2mm} $c$: Input conversation history (context);
    \STATE \hspace{2mm} $\mathcal{P}$: User-defined set of principles;
    \STATE \hspace{2mm} $\mathcal{T}$: Set of metrics corresponding to $\mathcal{P}$;
    \STATE \hspace{2mm} $i$: Initial generation prompt;
    \STATE \hspace{2mm} $r_p$: Refinement prompt for principle $p$;
    \STATE \hspace{2mm} $q$: Query generation prompt;
    \STATE \hspace{2mm} $u$: User utterance to make the response more $\{principle\}$;
    \STATE \hspace{2mm} $\tau = [\tau_1, \tau_2, \dots, \tau_k]$: Quality metric $i$ has threshold $\tau_i$;
    \STATE \hspace{2mm} $w = [w_1, w_2, \dots, w_k]$: Quality metric $i$ has weight $w_i$;
    \STATE \hspace{2mm} $\lambda$: Improvement threshold for weighted metric sum;
    \STATE \hspace{2mm} $N$: Number of initial responses generated per turn
    \STATE
    \STATE {\textbf{Until Max Turns Reached:}}
\end{algorithmic}
\begin{algorithmic}[1]
    \STATE $\mathcal{Q} \sim p_{\mathcal{M}}(y \hspace{1mm} | \hspace{1mm} d, c, q)$
    \STATE $c = c \cup \{\mathcal{Q}\}$
    \STATE $\mathcal{Y}_0 = \{y_n\}_{n=1}^N$ with $y_n \sim p_{\mathcal{M}}(y \mid d, c, i)$
    \STATE $y^0 = \argmax\limits_{y \in \mathcal{Y}_0}\left\{\sum\limits_{t=1}^{|\mathcal{T}|} \mathbbm{1}(\argmax\limits_{y' \in \mathcal{Y}_0}\{m_t(y', d, c)\} = y) \right\}$
    \IF{$\sum\limits_{t=1}^{|\mathcal{T}|} \mathbbm{1}(m_t(y^0, x)\geq \tau_t) = |\mathcal{T}|$}
        \STATE $c = c \cup \{y^0\}$
    \ELSE
        \FOR{iteration $j = 0, 1, \dots J$:}
            \STATE $\mathcal{Y}_{j+1} = \{y_p\}_{p=1}^{|\mathcal{P}|}$ with $y_p \sim p_{\mathcal{M}}(y \mid y^j, d, c, r_p)$
            \STATE $y^{j+1} = \argmax\limits_{y \in \mathcal{Y}_{j+1}}\left\{\sum\limits_{t=1}^{|\mathcal{T}|} \mathbbm{1}(\argmax\limits_{y' \in \mathcal{Y}_{j+1}}\{m_t(y', x)\} = y) \right\}$
            \footnotesize
            \IF{$\sum\limits_{t=1}^{|\mathcal{T}|} \mathbbm{1}(m_t(y^{j+1}, d, c)\geq \tau_p) = |\mathcal{T}|$}
                \STATE $c = c \cup \{y^j\} \cup \{u\} \cup \{y^{j+1}\}$
                \STATE \textbf{break}
            \ELSIF{$\sum\limits_{t=1}^{|\mathcal{T}|} w_t \cdot \mathbbm{1}(m_t(y^{j+1}, d, c) \geq m_t(y^{j}, d, c))$} %< \lambda$}
                \STATE $y^{j+1} = y^j$
            \ENDIF
        \ENDFOR
        \STATE $c = c \cup \{u\} \cup \{y^J\}$
    \ENDIF
\end{algorithmic}
\end{algorithm}
\end{minipage}
\end{figure}

\noindent We include a complete version of Algorithm 1 adapted for synthetic dialogue generation, leveraged in our fine-tuning experiments. At first, we sample a new user query from large language model $\mathcal{M}$, bootstrapping only off of the document. As the total number of turns (utterances) to be modeled is user-defined, we append each utterance to the end of the conversation history. For example, given the last user query, we append an agent response to it, which is either an initial (no refinement necessary) or final (post-refinement) response. If refinement did occur, then we first append the previous best agent response ($y^j$), then a user turn $u$ of "User: Please make this response more $\{principle\}$", and then the improved and sufficient agent response $y^{j+1}$. Once an agent response has been procured and appended for the current turn, we continue back to line 1 and generate a new user query, this time conditioning on the conversation history as well; this repeats until the user-specified max turn limit is reached. 

%\begin{figure}[h]
%    \centering
%    \includegraphics[scale=0.625]{self-refinement-algo-dialogue.png}
%    \caption{Extended description of the self-refinement algorithm in the setting of conversational question answering. Each component of the algorithm is explained in greater detail in Section 2, and the content-grounded question answering application domain is discussed in Section 3.}
%    \label{fig:enter-label}
%\end{figure}

\newpage
\section{Metric Selection and Threshold Calibration} \label{app:section-B}
\begin{table*}[h]
\label{sample-table}
\vskip 0.15in
\begin{center}
%\begin{scriptsize}
\begin{footnotesize}
\begin{sc}
\begin{tabular}{lccccccr}
\toprule
 Thresholding & Stage &  Rouge-L & BERT-Recall & BERT K-Prec. & Recall & K-Prec. \\
\midrule
% Rouge-1 Recall $\geq 0.02$ & Initial & 22.09 & 27.84 & 40.23 & 34.24 & 76.56 \\
%& Final & 22.68 & 29.73 & 43.75 & 36.88 & 79.88 \\
%\midrule
Rouge-1 K-Prec. $\geq 0.7$ & Initial & 22.06 & 27.60 & 39.45 & 34.1 & 76.08 \\
& Final & 22.21 & 28.19 & 40.29 & 34.57 & 77.23 \\
\midrule
 Rouge-1 K-Prec. $\geq 0.8$ & Initial & 23.06 & 28.65 & 39.65 & 35.42 & 76.49 \\
& Final & 23.26 & 29.36 & 41.00 & 36.34 & 76.68 \\
\midrule
Rouge-1 K-Prec. $\geq 0.9$ & Initial & 22.71 & 29.08 & 40.79 & 35.62 & 75.58 \\
& Final & 22.79 & 29.67 & 42.40 & 36.59 & 78.63 \\
%\midrule
%Rouge-1 Recall  & Initial & 22.70 & 28.30 & 40.06 & 34.81 & 76.75 \\
%& Final & 22.66 & 28.70 & 40.90 & 35.29 & 77.73 \\
\midrule
WeCheck $\geq 0.4$ & Initial & 24.00 & 29.87 & 44.02 & 36.45 & 81.51\\
 & Final & 23.93 & 30.02 & 44.53 & 36.67 & 82.08 \\
 \midrule
 WeCheck $\geq 0.5$ & Initial & 23.84 & 30.37 & 44.18 & 36.45 & 81.58\\
 & Final & 23.89 & 30.51 & 44.99 & 36.69 & 82.40 \\
 \midrule
 WeCheck $\geq 0.6$ & Initial & 24.37 & 30.19 & 44.27 & 36.48 & 81.31 \\
 & Final & 24.23 & 30.17 & 45.08 & 36.72 & 81.95 \\
 \midrule
 WeCheck $\geq 0.7$ & Initial & 23.85 & 30.05 & 44.48 & 36.27 & 81.75\\
 & Final & 23.82 & 30.38 & 45.64 & 37.00 & 82.63 \\
 \midrule
 WeCheck $\geq 0.8$ & Initial & 24.08 & 29.93 & 44.13 & 36.03 & 81.75\\
 & Final & 24.07 & 30.31 & 45.41 & 36.74 & 83.00 \\
 \midrule
 All 3 Rouge + WeCheck $\geq 0.4$   & Initial & 23.97 & 30.12 & 44.41 &  36.45 & 81.96    \\
& Final & 24.17 & 31.57 & 46.62 & 38.67 & 83.32 \\
\midrule
 All 3 Rouge + WeCheck $\geq 0.5$   & Initial & 24.02 & 30.08 & 44.28 &  36.86 & 81.29    \\
& Final & 24.05 & 31.29 & 46.40 & 38.74 & 82.81\\
\midrule
 All 3 Rouge + WeCheck $\geq 0.6$   & Initial & 24.01 & 29.91 & 44.55 &  36.07 & 81.50    \\
& Final & 23.90 & 31.11 & 46.80 & 38.05 & 83.26\\
\midrule
 All 3 Rouge + WeCheck $\geq 0.7$   & Initial & 23.90 & 30.06 & 44.02 &  36.49 & 81.49    \\
& Final & 24.17 & 31.62 & 46.22 & 38.88 & 83.00\\
\midrule
 All 3 Rouge + WeCheck $\geq 0.8$   & Initial & 23.79 & 29.94 & 44.47 &  36.49 & 81.84    \\
& Final & 23.35 & 31.08 & 47.00 & 38.37 & 83.57\\
\bottomrule
\end{tabular}
\end{sc}
%\end{scriptsize}
\end{footnotesize}
\end{center}
\vskip -0.1in
\vspace{1mm}
\caption{Threshold calibration and metric selection was performed on a development (validation) set split of the MultiDoc2Dial dataset, consisting of 2,038 samples. Experiments are reported with the Flan-T5-XXL \cite{flan-t5-palm} model. Note that Rouge-L between response and document (Rouge-L-Doc) as well as between response and the user query (Rouge-L-Query) are maintained constant, while we experiment with changing the third metric between Rouge-1 F1, Rouge-1 K-Precision, and Rouge-1 Recall. We also vary the threshold for the WeCheck reward model \cite{wecheck}, in isolation and in tandem with the best performing Rouge metric combination.}
\end{table*}

\noindent \paragraph{ROUGE Metric Thresholding.} We explore a plethora of different thresholding settings to calibrate sufficiency and select the metric set $\mathcal{T}$ accordingly. Note that "Rouge-1 K-Prec. $\geq 0.7$", refers to using Rouge-1 K-Precision with a threshold of 0.7, and Rouge-L F1 of the response with respect to both the document and the user query with a threshold of 0.05. Then, maintaining the latter two with the same configuration, we vary the first metric, exploring the use of a Rouge-1 K-Precision measure in place of recall and experimenting with thresholds of 0.7, 0.8, and 0.9. 

\noindent \paragraph{WeCheck and Combo Thresholding.} We also vary the threshold for the WeCheck reward model when applied as a standalone sufficiency metric, from 0.4 to 0.8 in increments of size 0.1. Finally, using the chosen combination of the three ROUGE metrics (i.e. Rouge-1 Recall with 0.02, Rouge-L F1 with 0.05 for both response-document and response-query comparison), we vary the WeCheck threshold, yielding a very interesting set of results. Across the majority of metrics, our self-refinement is effective for all thresholding methods applied, although it manifests to different degrees depending on the set of proxy metrics. Notably, the similarity in performance across threshold levels (i.e. not exhibiting a clear trend correlating to an increase in threshold) allows our algorithm to more effectively serve as a means of user-defined risk control with respect to a target refinement rate $\alpha$. As noted in Section 2, a greater threshold results in a higher standard for the initial response to meet, thus yielding a higher rate of refinement as more responses are deemed inadequate.

\section{Initial Response Generation, Query Generation, and Principle Refinement Prompts}
\label{app:section-C}
\subsection{Initial Generation Prompt}

\begin{figure}[h]
\begin{minipage}[h]{1\linewidth}
\begin{tcolorbox}[colback=gray!5,colframe=green!40!black]
\begin{verbatim}
Provided is a dialog between two speakers, User and Agent. Generate a response 
that is coherent with the dialog history and the provided document. Desired traits 
for responses are: 1) Specific - The response contains specific content, and 
2) Accurate - The response is correct and factual with respect to the document.

document: DIAL-IN search accounts#3_0Log On to DIAL - IN [1 ] \n\nWhat business 
records must I keep to document the searches I perform? \nThe business records you 
keep must exist prior to the search you perform and must establish the business 
purpose of the search. 
\end{verbatim}
\begin{center}
    \hspace{10mm} \vdots
\end{center}
\begin{verbatim}
To verify your browser is compatible to continue using any of the state's government 
websites, please visit https://encryption.ny.gov/ [6]. If your browser is not 
currently compatible , please update it to the newest version.

context: User: I need to know how to pay the dial-in search account fees.
Agent: The custoers must pay a deposit with the application, and it should be enough 
to pay for two months of searches. Was your application accepted?
User: Yes, it was.
Agent: then, your deposit will be added to your new account 
balance.
User: Can you tell me some of the organizations that are exempt from the search fees?
Agent: Some of the exempted organizations are any public organization, its officers, 
a volunteer fire company, volunteer ambulance service, etc. These organizations are 
exempt from the search fee.
User: What to do in case none of the users performed a search that the DMV contacted 
me for?
Agent: You should contact the DMV immediately.
User: Why would the DMV contact me about a search?
Agent: The DMV may contact you to ask you about a search to make sure you comply with
the Dial-In Terms of Service.

### 
\end{verbatim} \\
\begin{center}
    \hspace{10mm} \vdots
\end{center}
\end{tcolorbox} 
\end{minipage}
\end{figure}
\begin{figure}[t]
\begin{minipage}[h]{1\linewidth}
    \begin{tcolorbox}[colback=gray!5,colframe=green!40!black]
    \begin{center}
        \hspace{10mm} \vdots
    \end{center}
    \begin{verbatim} 
document: \n\nBenefits Planner: Family Benefits \nWhen you start receiving 
disability benefits , certain members of your family may also qualify for benefits 
on your record. Benefits may be paid to your : spouse; divorced spouse ; children; 
disabled child ; and adult child disabled before age 22.
\end{verbatim}
\begin{center}
    \hspace{10mm} \vdots
\end{center}
\begin{verbatim}
Find out more about Benefits For A Disabled Child. \n\nPublications \nDisability 
Benefits Benefits For Children What You Need To Know When You Get Social Security 
Disability Benefits Information for Government Employees Benefits For Children 
With Disabilities

context: User: Oh, hi. Please, i'm looking for some info about family benefits. 
could you help me out?
Agent: Are you currently receiving any disability benefits?
User: Yeah, i started to receive it recently.
Agent: Well, in that case, i can tell you that some members of your family may also 
qualify to get benefits on your record.

###

document: \n\nExposure through Project 112 or Project SHAD \nIf you were a part of 
chemical and biological warfare testing through Project 112 or Project Shipboard 
Hazard and Defense SHAD , you may be at risk for certain illnesses. The Department 
of Defense s Deseret Test Center in Fort Douglas, Utah, conducted this testing, 
which took place aboard ships and on land in various locations from 1962 to 1974. 
Find out if you can get disability compensation or benefits. \n\nCan I get 
disability benefits from VA? \nYou may be able to get disability benefits if you 
meet both of the requirements listed below.
\end{verbatim}
\begin{center}
    \hspace{10mm} \vdots
\end{center}
\begin{verbatim}
Get declassified Department of Defense fact sheets If you have a question about the 
tests , if you have any information that can help show you were part of them 
including whether you may have been part of them or contact the Department of 
Defense at 800 - 497 - 6261.

context: User: I wanted more information on VA benefits and project 112
Agent: Were you part of chemical and biological warfare testing through Project 112
or Project Shipboard Hazard and Defense SHAD?
\end{verbatim}
\end{tcolorbox}
\end{minipage}
\caption{Above is the initial generation prompt, containing the instruction and the three in-context exemplars drawn from the train set of MultiDoc2Dial \cite{multidoc2dial}, omitting the current sample inputs (document and context). The exemplars demonstrate question answering given the conversation history, and are separated by "\#\#\#". }
\end{figure}

\mbox{}
\clearpage
\newpage

\subsection{Query Generation Prompt for Synthetic Dialogue Generation}
%\vspace{-1mm}
\begin{figure}[h]
\begin{minipage}[h]{1\linewidth}
\begin{tcolorbox}[colback=gray!5,colframe=blue!40!black]
\begin{verbatim}
Provided is a dialog between two speakers, User and Agent. Generate a new question, 
posed by the user, that is coherent with the dialog history and contains 
specfic content.

document: \n\nBenefits Planner: Family Benefits \nWhen you start receiving 
disability benefits , certain members of your family may also qualify for benefits 
on your record. Benefits may be paid to your : spouse; divorced spouse ; children; 
disabled child ; and adult child disabled before age 22. 
\end{verbatim}
\begin{center}
    \hspace{10mm} \vdots
\end{center}
\begin{verbatim} 
Find out more about Benefits For A Disabled Child. \n\nPublications \nDisability 
Benefits Benefits For Children What You Need To Know When You Get Social Security 
Disability Benefits Information for Government Employees Benefits For Children 
With Disabilities 

context: User: Oh, hi. Please, i'm looking for some info about family benefits. 
could you help me out?
Agent: Are you currently receiving any disability benefits?
User: Yeah, i started to receive it recently.

###

document: NY State Adventure License FAQs#3_0\n\n7. Is there an additional fee to
have icons added to my DMV photo document? \nThere are no additional fees if you
request the icons be added at the time of your photo document renewal.
\end{verbatim}
\begin{center}
    \hspace{10mm} \vdots
\end{center}
\begin{verbatim}
For Boating Safety Certificate and Empire Passport holders , contact Parks via their
website: www.parks.ny.gov [3]. For Lifetime Sportsman, Small / Big Game, Bow 
Hunting, Trapping, Muzzle Loading, or Fishing, contact DEC via their website : 
www.dec.ny.gov [4].
\end{verbatim}
\begin{center}
    \hspace{10mm} \vdots
\end{center}
\end{tcolorbox}
\end{minipage}
\end{figure}
\begin{figure}[h]
\begin{minipage}[h]{1.05\linewidth}
\begin{tcolorbox}[colback=gray!5,colframe=blue!40!black]
\begin{center}
    \hspace{10mm} \vdots
\end{center}
\begin{verbatim}
context: User: I have a restricted use license issued in NJ and need information 
about driving in NY.
Agent: Do you meet NY requirements for obtaining a restricted license?
User: Yes, I do.
Agent: Great. You can receive a restricted license to drive in NY. The restrictions 
will be the same as the same as the restrictions for a driver with a NY driver license.
User: Where can I apply for the restricted driver license?

###

document: \n\nFamily Servicemembers Group Life Insurance (FSGLI) \nFamily SGLI, also 
known as Family Servicemembers Group Life Insurance FSGLI, offers coverage for the
spouse and dependent children of service members covered under full - time SGLI. 
\end{verbatim}
\begin{center}
    \hspace{10mm} \vdots
\end{center}
\begin{verbatim}
If your service member is part of the Public Health Service , you ll need to fill out the 
Spouse Coverage Election and Certificate SGLV 8286A and have them turn it in to their 
unit s personnel officer. Download the Spouse Coverage Election and Certificate PDF

context: User: How much will my service member pay for dependent coverage?
Agent: Nothing.s We provide dependent coverage at no cost until the child is 
18 years old , or sometimes longer if the child meets one of the requirements 
listed below
User: To continue receiving dependent coverage after age 18, what are the requirements? 
\end{verbatim}
\end{tcolorbox}
\end{minipage}
\caption{Query generation prompt (q in Appendix \ref{app:section-A}'s algorithm), containing an instruction and three in-context demonstrations of user queries given a document (separated by "\#\#\#"), omitting the current instance inputs. }
\end{figure}

\mbox{}
\newpage
%\clearpage

\subsection{Principle Refinement Prompt Template}

\begin{figure}[h]
\begin{minipage}[h]{1\linewidth}
\begin{tcolorbox}[colback=gray!5,colframe=teal!40!black]
\begin{verbatim}
document: {document}
        
context: {context}
Agent response 1 (not {principle}): {less_principle_response}
        
Let's make this response more {principle}. 
        
Agent response 2 (more {principle}): {more_principle_response}
\end{verbatim}
\end{tcolorbox}
\end{minipage}
\end{figure}

\mbox{}
\clearpage

\subsection{Specificity Principle Refinement Prompt with In-Context Exemplars}

\begin{figure}[h]
\begin{minipage}[h]{1.01\linewidth}
\begin{tcolorbox}[colback=gray!5,colframe=teal!40!black]
\begin{verbatim}
We want to improve the previous response to make it more specific. To aid in this
process, we provide examples of incremental improvement on specific, where Agent 
response 2 is more specific than Agent response 1.

document: DIAL-IN search accounts#3_0Log On to DIAL - IN [1 ] \n\nWhat business 
records must I keep to document the searches I perform? \nThe business records you 
keep must exist prior to the search you perform and must establish the business 
purpose of the search. 
\end{verbatim}
\begin{center}
    \hspace{10mm} \vdots
\end{center}
\begin{verbatim}
To verify your browser is compatible to continue using any of the state's government 
websites , please visit https://encryption.ny.gov/ [6]. If your browser is not 
currently compatible , please update it to the newest version.

context: User: I need to know how to pay the dial-in search account fees.
Agent: The custoers must pay a deposit with the application, and it should be enough 
to pay for two months of searches.
Agent: Was your application accepted?
User: Yes, it was.
Agent: then, your deposit will be added to your new account balance.
User: Can you tell me some of the organizations that are exempt from the search fees?
Agent: Some of the exempted organizations are any public organization, its officers,
a volunteer fire company, volunteer ambulance service, etc. These organizations are 
exempt from the search fee.
User: What to do in case none of the users performed a search that the DMV contacted 
me for?
Agent: You should contact the DMV immediately.
User: Why would the DMV contact me about a search?
Agent response 1 (not specific): The DMV may contact you about a search to ensure 
compliance.

Let's make this response more specific.

Agent response 2 (more specific): The DMV may contact you to ask you about a search to 
make sure you comply with the Dial-In Terms of Service.

###
\end{verbatim}
\begin{center}
    \hspace{10mm} \vdots
\end{center}
\end{tcolorbox}
\end{minipage}
\end{figure}

\begin{figure}[h]
\begin{minipage}[h]{1.01\linewidth}
\begin{tcolorbox}[colback=gray!5,colframe=teal!40!black]
\begin{center}
    \hspace{10mm} \vdots
\end{center}
\begin{verbatim}
document: \n\nBenefits Planner: Family Benefits \nWhen you start receiving 
disability benefits , certain members of your family may also qualify for benefits 
on your record. Benefits may be paid to your : spouse; divorced spouse ; children; 
disabled child ; and adult child disabled before age 22. 

context: User: Oh, hi. Please, i'm looking for some info about family benefits. 
could you help me out?
Agent: Are you currently receiving any disability benefits?
User: Yeah, i started to receive it recently.
Agent response 1 (not specific): Then, some others may qualify for benefits.

Let's make this response more specific. 

Agent response 2 (more specific): Well, in that case, i can tell you that some 
members of your family may also qualify to get benefits on your record.

###

document: \n\nExposure through Project 112 or Project SHAD \nIf you were a part of 
chemical and biological warfare testing through Project 112 or Project Shipboard
Hazard and Defense SHAD , you may be at risk for certain illnesses.
\end{verbatim}
\begin{center}
    \hspace{10mm} \vdots
\end{center}
\begin{verbatim}
Get declassified Department of Defense fact sheets If you have a question about the 
tests , if you have any information that can help show you were part of them including 
whether you may have been part of them or contact the Department of Defense at 
800 - 497 - 6261.

context: User: I wanted more information on VA benefits and project 112
Agent response 1 (not specific): Were you part of Project 112 or Project SHAD?

Let's make this response more specific.

Agent response 2 (more specific): Were you part of chemical and biological warfare 
testing through Project 112 or Project Shipboard Hazard and Defense SHAD?
\end{verbatim}
\end{tcolorbox}
\end{minipage}
\caption{Refinement prompt ($r_p$) with respect to the specificity principle, with an instruction and three in-context demonstrations. Our instruction explicitly suggests to improve on specificity, and that in the provided in-context exemplars, the latter response (Agent response 2) is a specificity improvement over the former response (Agent response 1). Notably, we demonstrate to the model that "Let's make this response more specific" is an utterance in between the worse and better responses. Each exemplar and the "more specific" (gold) response is derived from the MultiDoc2Dial train set, while the "not specific" response is developed by a human annotator, bootstrapping off the gold response. The three exemplars are separated by "\#\#\#", and the above prompt omits the current instance inputs.}
\end{figure}

\mbox{}
\clearpage

%\section{Automated In-Context Exemplar Generation}
%With the aim to determine if task adaptation can be automated aside from metric selection and threshold calibration, we attempt to worsen the gold response along each principle to yield a refinement delta without manual annotation. We provide the prompt “make your response less \{principle\},” and evaluate in both 0-shot and 3-shot settings. Surprisingly, worsening the response along the specificity dimension also makes it noticeably less faithful in the process, making the response less relevant instead worsens all three dimensions, and lessening faithfulness yields minimal change. Given our results demonstrate that just three in-context exemplars are sufficient, it thus far appears that it is preferable to manually curate a small number of demonstrations for refinement rather than rely on generating them. We suggest this task as a valuable area for further exploration in studying and reducing the intertwinement between principles, such that one could improve or worsen responses in a dimension-specific manner as a result of improvements in instruction-following capabilities. 

%\newpage
\section{Fine-tuning Experimental Setup}
\label{sec:fine-tuning}

We QLoRA fine-tune \textsc{llama-2-13b-chat} on both human annotated and synthetically generated MultiDoc2Dial datasets. We set the learning rate to 1e-5, LoRA rank to 8 and LoRA alpha to 32. We apply 4bit quantization for both model training and inferencing. Unlike baseline model inferencing with few-shot learning for which we use sampling method, we use greedy decoding for fine-tuned models.

We train the models with 4 A100 (80GB memory) GPUs up to 10 epochs. Training takes between 5 hours for 8k training samples and 24 hours for about 50k samples. We select the best checkpoint on the basis of the 5 evaluation metrics (RougeL, BERTScore Recall, BERTScore K-Prec., Recall and K-Prec.) scores on the development test data. 

%\section{Evaluation Datasets}
%\newpage

\section{Zero-Shot vs Few-Shot Comparison}
\begin{table*}[h]
%\vskip 0.15in
\begin{center}
%\begin{scriptsize}
%\begin{footnotesize}
\begin{footnotescript}
\begin{sc}
\begin{tabular}{lccccccr}
\toprule
 Initial Exemplars + Metrics & Stage & Rouge-L & BERT-Recall & BERT K-Prec. & Recall & K-Prec. \\
\midrule
 Flan-T5-XXL (11B) Results & & & & & & \\
\midrule
 3-Shot / Only Rou-L + Rou-1 & Initial & 21.50 & 27.27 & 38.08 & 31.19 & 75.58 \\
 & Final & 21.84 & 28.96 & 41.41 & 33.96 & 78.78 \\
\hline
 0-Shot / Only Rou-L + Rou-1 & Initial & 21.55 & 28.11 & 40.42 & 32.34 & 76.77 \\
& Final & 21.72 & 29.29 & 42.74 & 34.14 & 79.29 \\
\hline
 3-Shot / Only RM & Initial & 22.67 & 28.80 & 42.83 & 33.00 & 80.42\\
 & Final & 22.65 & 28.91 & 43.55 & 33.31 & 81.14 \\
 \hline
  0-Shot / Only RM & Initial & 22.33 & 28.91 & 44.58 & 33.61 & 81.29\\
 & Final & 22.43 & 29.17 & 45.60 & 34.20 & 82.25 \\
 \hline
 3-Shot / Rou + RM    & Initial & 22.61 & 28.75 & 42.60 &  32.75 & 80.36    \\
 & Final & \textbf{22.71} & 29.89 & 44.86 & 34.76 & 81.98\\
\hline
 0-Shot / Rou + RM   & Initial & 22.30 & 28.94 & 44.55 &  33.68 & 81.56    \\
& Final & 22.38 & \textbf{30.10} & \textbf{46.60} & \textbf{35.58} & \textbf{83.13}\\
\hline\midrule
Llama-2-13B-Chat Results & & & & & & \\
\midrule
 3-Shot / Only Rou-L + Rou-1 & Initial & 20.63 & 29.35 & 34.32 & 36.49 & 71.03 \\
& Final & 20.23 & 30.22 & 36.07 & 38.82 & 72.74        \\
\hline
 0-Shot / Only Rou-L + Rou-1 & Initial & 19.31 & 28.92 & 34.44 & 38.45 & 70.33 \\
 & Final & 18.95 & 29.67 & 36.04 & 40.43 & 71.76        \\
\hline
 3-Shot / Only RM & Initial & \textbf{21.50} & 30.35 & 39.20 & 36.89 & 76.29 \\
& Final & 21.25 & 30.11 & 39.51 & 36.85 & 76.53 \\
\hline
 0-Shot / Only RM & Initial & 19.97 & 29.65 & 34.33 & 38.07 & 70.08 \\
& Final & 19.95 & 29.89 & 40.68 & 38.59 & 76.73 \\
\hline
 3-Shot / Rou + RM & Initial & 21.48 & 30.29 & 39.15 & 36.79 & 76.34 \\
& Final & 21.11 & \textbf{30.95} & 40.05 & 38.62 & 76.89 \\
\hline
 0-Shot / Rou + RM & Initial & 20.36 & 30.17 & 40.68 & 38.59 & 76.84 \\
& Final & 20.06 & 30.64 & \textbf{41.46} & \textbf{40.00} & \textbf{77.43} \\
\bottomrule
\end{tabular}
\end{sc}
%\end{scriptsize}    
%\end{footnotesize}
\end{footnotescript}
\end{center}
\vskip -0.1in
\caption{Experimental Results on MultiDoc2Dial test set, containing 10,204 instances. Experiments are reported with the Flan-T5-XXL and Llama-2-13B-Chat models, using 3 Rouge (ROU) measures, the WeCheck reward model (abbreviated as RM), and both in tandem for sufficiency thresholding. Highest scores are boldfaced for each model. The zero-shot results are the same as those included in Table \ref{tab:combined_inference}; this table presents a comparison between 0-shot and 3-shot performance, for each metric set. 
}
\label{tab:zero-few-shot}
\end{table*}

\noindent In Table \ref{tab:zero-few-shot}, we also present a comparison based on the number of few-shot exemplars employed in initial response generation. Notably, we observe that the 0-shot performance of initial responses are, in fact, higher than the 3-shot results for the same phase.  %but the 1-shot and 3-shot results are remarkably similar.
This suggests that instruction-tuned LMs such as Flan-T5-XXL are already fairly adept at dialogue response generation without in-context exemplars. Furthermore, we find that the zero-shot setting achieves higher initial scores relative to 3-shot for Flan-T5-XXL, but such improvement is less consistent for Llama-2-13B-Chat. Simultaneously, we found that the 3-shot results with refinement constitute an improvement over the 0-shot performance. This indicates that in-context exemplars are necessary to improve performance during the refinement phase, although only three demonstrations are sufficient to illustrate the notion of the target principle to the LM. That is, despite fairly coherent initial responses, there is still room for improvement, achieved when using three in-context exemplars per principle. Thus, the results reported above and in Table \ref{tab:combined_inference} hold 3 exemplars constant for the refinement phase. 

\newpage
\section{LLM-as-a-Judge Evaluation Setup} \label{app:section-F}

Recent literature \cite{zheng2023judging, zhang2024llmjudge} evidences the ability of using language models as discriminators, judging generation quality in lieu of (or as a supplement to) human evaluation feedback. The line of work has also been a strong motivator in influencing self-feedback and refinement approaches; it demonstrates the ability of powerful models to reflect human preferences and provide meaningful critiques. In pursuing such an approach, we deliberately choose to explicitly model certain properties in the instruction: for example, we seek permutation-invariance (while knowing that models are susceptible to position bias when given a set of multiple choice options and mitigating their preference for longer answers.

\subsection{Judge Prompt Template}

\begin{figure}[h]
\begin{minipage}[h]{1.05\linewidth}
\begin{tcolorbox}[colback=gray!5,colframe=olive!40!black]
\begin{verbatim}
Please act as an impartial judge and evaluate the quality of the responses provided by 
the two AI assistants to the user question displayed below. Your evaluation should 
consider correctness and helpfulness. You will be given a reference document, a user 
conversation, assistant A's answer, and assistant B's answer. Your job is to evaluate 
which assistant's answer is better based on the information in the reference document 
and the user conversation so far. Begin your evaluation by comparing both assistants' 
answers with the document and the user conversation so far. Identify and correct any
mistakes. Avoid any position biases and ensure that the order in which the responses 
were presented does not influence your decision. Do not allow the length of the 
responses to influence your evaluation. Do not favor certain names of the assistants. 
Be as objective as possible. After providing your explanation, output your final verdict 
by strictly following this format: "[[A]]" if assistant A is better, "[[B]]" if 
assistant B is better, and "[[C]]" for a tie.

[User Document]
{document}

[User Conversation]
{conversation}

[The Start of Assistant A's Answer]
{answer_a}
[The End of Assistant A's Answer]

[The Start of Assistant B's Answer]
{answer_b}
[The End of Assistant B's Answer]
\end{verbatim}
\end{tcolorbox}
\end{minipage}
\caption{LLM-as-a-Judge prompt template for automated response evaluation between initial and final generations in the content-grounded question answering setting. A document and conversation are given as an input for each sample, and we compare two possible agent responses to the most recent user query posed in the conversation.  }
\end{figure}

%%%%%%%%%%%%%%%%%%%%%%%%%%%%%%%%%%%%%%%%%%%%%%%%%%%%%%%%%%%%%%%%%%%%%%%%%%%%%%%
%%%%%%%%%%%%%%%%%%%%%%%%%%%%%%%%%%%%%%%%%%%%%%%%%%%%%%%%%%%%%%%%%%%%%%%%%%%%%%%

\end{document}